\documentclass[runningheads]{llncs}
\usepackage{graphicx}
\usepackage{comment}
\usepackage{amsmath,amssymb} 
\usepackage{color}
\usepackage{array}
\usepackage{multirow}
\usepackage{subcaption}
\usepackage{booktabs}
\usepackage[hidelinks]{hyperref}
\usepackage{pifont}
\newcommand{\cmark}{\text{\ding{51}}}
\newcommand{\xmark}{\text{\ding{55}}}
\usepackage[misc]{ifsym}

\makeatletter
\newcommand*\fsize{\dimexpr\f@size pt\relax}
\makeatother
\renewcommand{\orcidID}[1]{\href{https://orcid.org/#1}{\includegraphics[height=1\fsize]{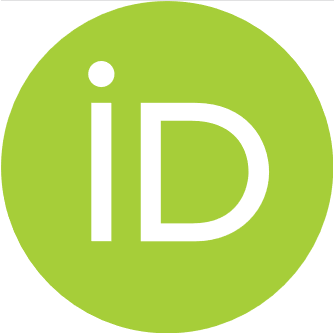}}}

\begin{document}
\pagestyle{headings}
\mainmatter
\def\CVPPPSubNumber{X}  
\def\ECCVSubNumber{\CVPPPSubNumber} 
\title{Object Detection Based Handwriting~Localization} 


\titlerunning{Object Detection Based Handwriting Localization}

\author{Yuli Wu \inst{1}\thanks{This work was done during their internship at SAP ICN Nanjing.} \orcidID{0000-0002-6216-4911} 
	\and Yucheng Hu\inst{2}\textsuperscript{\thefootnote} \orcidID{0000-0002-6321-4837}
	\and Suting Miao\inst{3} \href{mailto:phoebe.miao@sap.com}{\Letter} }
\authorrunning{Yuli Wu et al.}

\institute{Rheinisch-Westfälische Technische Hochschule Aachen, Germany \\
\and
Nanjing Normal University, China \\
\and
SAP Innovation Center Network (ICN) Nanjing, China}

\maketitle
\begin{abstract}
We present an object detection based approach to localize handwritten regions from documents, which initially aims to enhance the anonymization during the data transmission. The concatenated fusion of original and preprocessed images containing both printed texts and handwritten notes or signatures are fed into the convolutional neural network, where the bounding boxes are learned to detect the handwriting. Afterwards, the handwritten regions 
can be processed (\textit{e.g.} replaced with redacted signatures) to conceal the \textit{personally identifiable information} (PII). This processing pipeline based on the deep learning network Cascade R-CNN works at 10 fps on a GPU during the inference, which ensures the enhanced anonymization with minimal computational overheads. Furthermore, the impressive generalizability has been empirically showcased: the trained model based on the English-dominant dataset works well on the fictitious unseen invoices, even in Chinese. The proposed approach is also expected to facilitate other tasks such as handwriting recognition and signature verification.

\end{abstract}

\setcounter{footnote}{0} 
\section{Introduction}
Handwriting localization plays an important role in the following scenarios: First, the handwritten regions in the documents may contain sensitive information, which must be anonymized before transmission. Second, handwriting localization can be naturally served as the first stage to achieve \textit{handwriting-to-text} recognition. Third, signatures to be verified must be extracted via localization from their surrounding texts or lines in the documents.

This work is initially motivated by the demanding case of anonymization enhancement. Access to data is vital to undertake enterprise today. One of the most common data types would be the invoices: with digitalized invoices and all kinds of powerful AI-driven technologies, the companies would be able to analyze customers’ behaviors and extract business intelligence automatically, offering utmost help to refine strategies and make decisions. However, the \textit{personally identifiable information} (PII) must be anonymized beforehand, as it is not worth the risk of any privacy exposure.

Ostensibly, tabular texts are of overwhelming majority in the business processing. In fact, the documents containing handwritten notes or signatures, such as invoices, also play an important role. The goal of this work is to localize the handwritten regions from the full-page invoice images for anonymization enhancement. The detected handwriting shall be anonymized afterwards, while the detailed implementation of which is beyond the scope of this work. As the expected results of the whole processing, handwriting should be excluded from the anonymized invoices, where it is assumed all handwritten regions would contain PII. One trivial way to realize this would be replacing the handwritten boxes with redacted signatures or notes. 

Tesseract~\cite{smith2007overview}, the de facto paradigm of \textit{optical character recognition} (OCR) engines, is nowadays widely used in the industry to extract textual information from images. OCR engines are competent to deal with the \textit{optimal} data, which is referred to as the image data of documents, where all items of interest are regularly printed texts under the context of this work. In contrast, the real-world documents are usually the ones containing not only the regularly printed texts, but also some irregular patterns, such as handwritten notes, signatures, logos etc, which might also be desired.

In this work, we adopt object detection approaches with deep learning networks to localize handwritten regions in the document data based on the SAP's Data Anonymization Challenge\footnote{\url{https://www.herox.com/SAPAI/}}. The feasibility and effectiveness of such algorithms have been empirically shown on those scenarios, where the \textit{objects} (handwriting) and the \textit{backgrounds} (printed texts) are extremely similar. Besides, the improvement from Faster R-CNN~\cite{ren2016faster} to Cascade R-CNN~\cite{cai2019cascade} can be effortlessly reproduced. In addition, the new baseline of the handwriting localization as the subtask from the SAP's Data Anonymization Challenge\footnotemark[1] has been released. Last but not least,
the proposed deep learning approach with Cascade R-CNN~\cite{cai2019cascade} has demonstrated impressive generalizability. The trained model based on the English-dominant dataset works well on the fictitious unseen invoices, even for those in Chinese as toy examples. Empirically, it is believed that the deep learning model has learned the \textit{irregularity} of the images.

Since the detailed types of handwritten regions, such as signatures or notes, are not discriminated during the experiments, we term \textit{detection} and \textit{localization} in this work interchangeably. The one-class detection merely consists of the localization regression task without classification. Despite the simplicity of the task description, it is still challenging to distinguish the handwritten notes from the printed texts, as they are similar regarding the contextual information. Furthermore, the detected bounding boxes, which should contain PII, are expected to be more accurate, compared to the general object detection tasks, \textit{i.e.}, the primary evaluation score \(AP^{\,FP}\) (average precision with  penalty of false positive, see Section~\ref{sec:eval}) is thresholded with the IoU of 80\%.

\clearpage
\section{Related Work}
The input images are usually in the format of the cropped handwritten regions in the signature verification competitions~\cite{malik2013icdar,ortega2003mcyt}. Likewise, some handwritten text recognition datasets provide the option of the images labeled with divided lines~\cite{marti2002iam}. This work is expected to bridge the gap between these researches and industrial applications through handwriting localization. 
Besides, text detection in natural scene images is close to our work. One significant difference between these two tasks is the \textit{target objects}: All texts should be detected in scene text detection task (\textit{e.g.}~\cite{nayef2019icdar2019}), while only the handwritten texts in this work. The other difference is the background: The background in scene text detection task is the natural view. In this work, the background is the printed texts and tables on the blank document. Also based on Faster R-CNN~\cite{ren2016faster}, Zhong et al.~\cite{zhong2017improved} uses LocNet~\cite{gidaris2016locnet} to improve the accuracy in scene text detection, whereas we use Cascade R-CNN~\cite{cai2019cascade}, the cascade version of Faster R-CNN.

There are two main categories of methods to localize the handwritten regions in the documents. The OCR based approaches recognize and then exclude printed texts. As a result, the unrecognizable parts are believed to be the handwriting. In contrast, the object detection based approaches regard this as a localization task, where the handwriting is the target and all other items (such as printed texts, logos, tables, etc.) are considered as the background. Thanks to the datasets and detection challenges on common objects (\textit{e.g.}~\cite{voc,lin2014coco}), a considerable number of novel algorithms about object detection have been productively proposed in the recent years, \textit{e.g.} Faster R-CNN~\cite{ren2016faster}, YOLO~\cite{redmon2016you}, SSD~\cite{liu2016ssd}, RetinaNet~\cite{lin2017focal}, Cascade R-CNN~\cite{cai2019cascade}, etc.

Three different approaches submitted to the Data Anonymization Challenge\footnotemark[1] are also briefly introduced in the following sections, including an OCR based approach and two deep learning based approaches (one with YOLOv3~\cite{redmon2018yolov3}, one with Google's paid cloud service).

\textbf{OCR based approaches.} 
In this section, an example proposal from the challenge\footnotemark[1] is demonstrated. First, the images are sequentially preprocessed, including removing the horizontal and vertical lines, median filtering (to remove salt-and-pepper noises), thresholding and morphological filtering (\textit{e.g.} dilation and erosion). The handwritten parts are then discriminated from the printed ones with respect to the manually chosen features like the heights and widths of the text boxes, text contents and confidence scores recognized by OCR. In the experiments, this approach brings in the results on a par with those using deep learning approaches. However, the robustness and the generalizability of the deep learning approaches are believed to be advantageous.

\textbf{Object detection based approaches with deep learning.} Since object detection is an intensively researched area in the field of computer vision, it is natural to directly apply the deep learning algorithms to the handwriting localization task. With the deep learning engine ImageAI~\cite{ImageAI}, the networks like YOLOv3~\cite{redmon2018yolov3} can be trained in an end-to-end manner. Moreover, some deep learning services like Google's Cloud AutoML Vision API take it further, managing the training process even without specifically assigning an algorithm.

\clearpage

\begin{figure}[t]
\centering
	\begin{subfigure}{0.33\textwidth}
	    \centering
        \includegraphics[height=4.2cm]{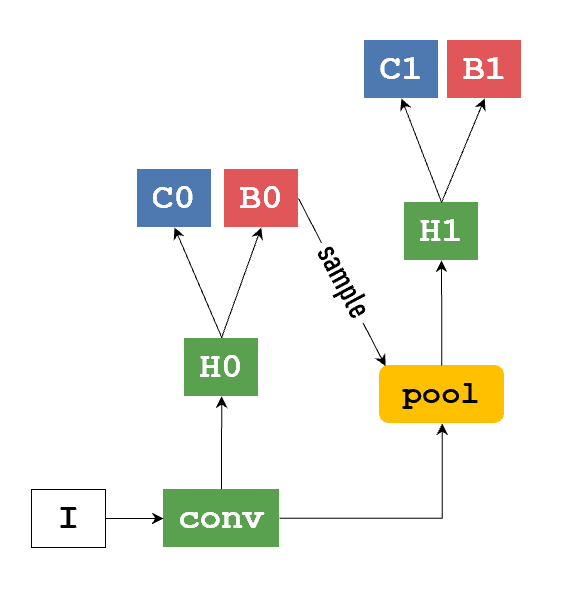}
    	\caption{Faster R-CNN}
    	\label{fig:faster}
	\end{subfigure}
	\begin{subfigure}{0.59\textwidth}
	\centering
    	\includegraphics[height=4.2cm]{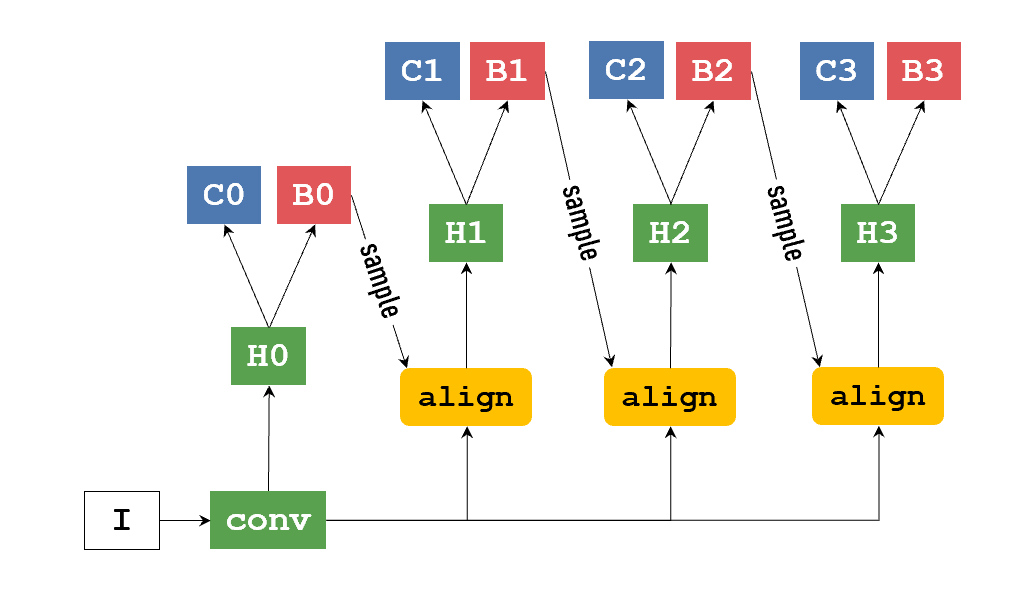}
    	\caption{Cascade R-CNN}
    	\label{fig:cascade}	
	\end{subfigure}
	\caption{Network Architectures of Faster R-CNN~\cite{ren2016faster} and Cascade R-CNN~\cite{cai2019cascade}. Figures are adapted from~\cite{cai2019cascade}.}
	\label{fig:net}
\end{figure}

\section{Method}\label{sec:method}

\subsection{Faster R-CNN}
Faster R-CNN~\cite{ren2016faster} consists of two modules: the Region Proposal Network (RPN) that proposes rectangular regions containing the desired objects, and the Fast R-CNN detector~\cite{girshick2015fast} that predicts the classes and the locations. 

The processing pipeline is demonstrated based on Fig.~\ref{fig:faster}. The input images (\texttt{I}) are first fed into a convolutional neural network (\texttt{conv}), where the shared features are extracted for both RPN and Fast R-CNN detector. Given the shared convolutional feature map of a size \(w\times h\times d\) and the number of the anchors \(k\) for each location in the feature map, the RPN head (\texttt{H0}) transforms it into two proposal features of \(w\times h\times 2k\) (\texttt{C0}) and \(w\times h\times 4k\) (\texttt{B0}) with one \textit{e.g.} \(3\times 3\) convolutional layer followed by two sibling \(1\times 1\) convolutional layers. 

Now, \(w\times h\times k\) proposals have been generated, each in the form of 6 representative values: 2 objectness scores and 4 coordinate offsets. The higher-scored proposals from \texttt{B0} are selected as the inputs of the Fast R-CNN detector, together with the shared convolutional feature map. The transform of the proposals' coordinates between the original images and the feature maps is calculated via \textit{e.g.} RoIPool~\cite{girshick2015fast} (\texttt{pool}) or RoIAlign~\cite{he2017mask} (\texttt{align}). The pooled or aligned \textit{region of interest} (RoI) feature map of some fixed size is flattened then projected onto a feature vector via the RoI head (\texttt{H1}). Finally, two vectors of classes (\texttt{C1}) and locations (\texttt{B1}) are obtained by fully connected layers upon the feature vector.

There are two places where multi-task loss functions are calculated: RPN (\texttt{C0} and \texttt{B0}) and Fast R-CNN detector (\texttt{C1} and \texttt{B1}). First, log loss is used for both classification tasks (specifically, sigmoid activation function plus binary cross entropy loss for \texttt{C0} and softmax activation function plus cross entropy loss for \texttt{C1}). Second, smooth L1 loss~\cite{girshick2015fast} is used for both bounding box regression tasks (\texttt{B0} and \texttt{B1}), which is defined as: \(\mathrm{smooth}_{L1}(x)= 0.5x^2\) if \(|x|<1\) and \(\mathrm{smooth}_{L1}(x)= |x|-0.5\) otherwise.

\clearpage
\subsection{Cascade R-CNN}
Fig.~\ref{fig:net} (adapted from~\cite{cai2019cascade}) depicts the differences of these two framework architectures. First, a cascade network is used to train the regressors and classifiers in a multi-stage manner. A four-stage version is illustrated in Fig.~\ref{fig:cascade}, including one RPN stage and three detection stages. Second, the IoU threshold is different for each detection stage, which is increasingly set to \(\{0.5,\,0.6,\,0.7\}\). Note that the mentioned IoU threshold does \textit{not} refer to the one in the RPN or the one when calculating \(mAP\) (mean Average Precision~\cite{lin2014coco}). It is used to define the positive or negative candidates during the mini-batch sampling (\texttt{sample} in Fig.~\ref{fig:net}).

Thanks to the cascade architecture with progressively increasing IoU thresholds for sampling, Cascade R-CNN can accomplish object detection of \textit{high quality}, which is exactly desired in the handwriting localization task to enhance anonymization.

\subsection{Other Techniques}
\textbf{\(\text{Canny Edge Detection}\).} The gradient intensity based Canny edge detector~\cite{canny1986computational} generates preliminary edges for the following processes. The detected edges might be truncated, \textit{e.g.} under different optical circumstances. Thus, further processes of refinement or extraction are normally applied to the edges detected by Canny.

\noindent
\textbf{Hough Transform.} Through the transform of line parameterizations, straight lines can be efficiently detected by the voting based Hough Transform~\cite{duda1972hough}. The images are usually first processed by edge detectors like Canny, followed by thresholding. Next, each edge pixel (\(x,y\)) in the binary images is represented by \(k\) evenly rotating lines through it in the Hesse normal form: \(r=x\cos(\theta)+y\cos(\theta)\). In the so-called \textit{accumulator space} of (\(r,\theta\)), each edge pixel would have \(k\) votes. The peaks of the accumulator space are thus the desired lines. In this work, lines detected by Hough Transform are removed in the preprocessing step to generate clearer input images for the deep learning network.

\noindent
\textbf{Tesseract OCR Engine.} As an open source paradigm OCR engine, Tesseract~\cite{smith2007overview} has been widely used to recognize textual information in the industry. In this work, a Python wrapper for the tesseract-ocr API has been used (\url{https://github.com/sirfz/tesserocr}) to detect and eliminate the printed texts in the preprocessing step.

\section{Experiments}\label{sec:exp}
\subsection{Dataset}\label{sec:dataset}
The dataset used in this work is the scanned full-page low-quality invoices in the tobacco industry from the 1990s (\url{http://legacy.library.ucsf.edu/}), which was once served in a document classification challenge~\cite{harley2015evaluation}. Based on the invoice (or invoice-like) images from the same dataset, the labels and the bounding boxes of names and handwritten notes are manually annotated for the Data Anonymization Challenge\footnotemark[1]. 

In total, we have access to 998 gray-scale images with ground-truth labels, which are randomly split to 600, 198 and 200 images as training, validation and testing set, respectively (denoted below as \texttt{600train+198val+200test}). The hidden evaluation set consists of 400 images. In this case the training set covers 800 images, the validation set remains unchanged and the testing set covers 400 unseen images (denoted below as \texttt{800train+198val+400test}). The sizes of the images are varied around 700\(\times\)1000, which are resized to 768\(\times\)768.

\subsection{Preprocessing}\label{sec:prep}
Despite the powerful capability of extracting features \textit{automatically} being one of the benefits when using deep learning algorithms, it is believed that the elementarily preprocessed inputs or their fusions might improve the performance. Intuitively, if some non-handwritten parts could be omitted in the preprocessing step, the following handwriting localization task could be facilitated. Based on this assumption, texts recognized by the OCR engine (\texttt{tesseract-ocr}~\cite{smith2007overview}) of high confidence and the straight lines detected by Hough transform~\cite{duda1972hough} are excluded. In the experiments, the threshold confidence for the OCR engine is set to 0.7. The preprocessed images without highly confident textual information or straight lines of tables are denoted as \texttt{"pre"} in the following, while the original ones as \texttt{"o"}.

Besides, the documents usually consist of a white background (of the highest intensity values, \textit{e.g.} 1) and black texts (of the lowest intensity values, \textit{e.g.} 0). As the background is dominant in terms of the number of pixels, it is natural to negate the images to obtain the inputs of sparse tensors, which is believed to make the learning progress more effectively. Given an image ranged in \([0,1]\), the negated image is calculated as the original image element-wise subtracted by 1. With this in mind, the negated original and preprocessed images are denoted as \texttt{"o-"} and \texttt{"pre-"}, respectively. 

In addition, the inputs of the deep learning networks can be usually of an arbitrary number of channels. The original and preprocessed images are concatenated to create fused inputs, where the preprocessed layer can be interpreted as an \textit{attention} mechanism, which highlights the most likely regions being the target objects. The concatenated inputs are denoted as \textit{e.g.} \texttt{"o/pre"}, the two dimensions of which are original and preprocessed images. The influences of the different inputs are investigated in Section~\ref{sec:res}.

\subsection{Training with Deep Learning Networks}\label{sec:dlnetwork}
All experiments were running on a single RTX 2080 Ti GPU. The implementation of the deep learning networks are adopted by the open source toolbox from OpenMMLab for object detection: MMDetection~\cite{mmdetection}. 

The default optimizer is \textit{Stochastic Gradient Descent} (SGD~\cite{lecun1989backpropagation}) with a learning rate of 0.001, a momentum of 0.9, and a weight decay of 0.0001. Since the training set of 600 images is relatively small, the default number of epochs is set to a relatively large one (200) to make full use of the computational capacity if the training lasts overnight. During the experiments, it is observed that 200 epochs are appropriate (Fig.~\ref{fig:epoch}). The \texttt{train/val/test} sets are defined as in Section~\ref{sec:dataset}. Model weights after each epoch with the best results of \texttt{val} set are chosen to make predictions on \texttt{test} set. The different preprocessing steps are introduced in Section~\ref{sec:prep} and they are compared with Faster R-CNN~\cite{ren2016faster} and Cascade R-CNN~\cite{cai2019cascade} in details. Next, additional two deep learning networks, RetinaNet~\cite{lin2017focal} and YOLOv3~\cite{redmon2018yolov3}, have been tested with the preprocessing step which yields the best result on Cascade R-CNN. Detailed experimental results can be found in Section~\ref{sec:res}.

\subsection{Evaluation Scores}\label{sec:eval}
\subsubsection*{\(\mathbf{IoU}\).} Intersection over Union of two bounding boxes \(p\) and \(g\) is defined as below:
\begin{equation}
    \mathrm{IoU}(p,g) = \frac{\mathrm{Area}\{\,p\cap g\,\}}{\mathrm{Area}\{\,p\cup g\,\}}
\end{equation}

\subsubsection*{\(\mathbf{GIoU}\).} Global IoU of two lists of bounding boxes \(P=\{p_1,p_2,...\}\) and \(G=\{g_1,g_2,...\}\) is defined as below:
\begin{equation}
    \mathrm{GIoU}(P,G) = \frac{\mathrm{Area}\{\,(p_1\cap p_2\cap ...)\cap (g_1\cap g_2\cap ...)\,\}}{\mathrm{Area}\{\,(p_1\cap p_2\cap ...)\cup (g_1\cap g_2\cap ...)\,\}}
\end{equation}

\subsubsection*{\(\mathbf{AP^{FP}}\).} Average Precision with penalty of False Positive is the original evaluation score for the handwriting detection used in the Data Anonymization Challenge\footnotemark[1], which is defined as follows:
\begin{equation}
    AP^{FP}= \begin{cases}\displaystyle
     \frac{|\mathcal{M}^G|}{|\mathcal{G}|} \cdot 0.75^{|\mathcal{P}|-|\mathcal{M}^P|} ,\; \mathrm{if}\; |\mathcal{G}| \ne 0;\vspace{0.2cm}\\ \; 0.75^{|\mathcal{P}|-|\mathcal{M}^P|}, \; \mathrm{otherwise}.
    \end{cases}
    \label{eq:score}
\end{equation}

In Eq.~\ref{eq:score}, \(\mathcal{P},\; \mathcal{G},\; \mathcal{M}^G,\; \mathcal{M}^P\) denote the sets of predicted, ground-truth, matched \textit{w.r.t.} ground-truth and matched \textit{w.r.t.} predicted bounding boxes, respectively, and \(|\cdot|\) denotes the number of the bounding boxes in this set. 

The criterion to call some predicted bounding box \(p_i\in\mathcal{P}\) a match \textit{w.r.t.} the ground-truth bounding box \(g\in\mathcal{G}\) is, when the IoU between \(p_i\) and \(g\) is greater than a threshold \(T\), \textit{i.e.}:
\begin{equation}
\mathcal{M}^G = \{\, g\in\mathcal{G} \;|\; \exists\;p_i\in\mathcal{P}\, \big(\, \mathrm{IoU}(p_i\,,\, g)>T\,\big)\}.
\end{equation}

Analogously, 
\begin{equation}
\mathcal{M}^P = \{\, p\in\mathcal{P} \;|\; \exists\;g_i\in\mathcal{G}\, \big(\, \mathrm{IoU}(p\,,\, g_i)>T\,\big)\}.
\end{equation}

It differs from the popular evaluation score  AP (Average Precision) for object detection in COCO~\cite{lin2014coco}, where the false positive (FP) has not been particularly punished. Moreover, considering the potential application on the anonymization enhancement, the IoU threshold \(T\) in the Data Anonymization Challenge\footnotemark[1] is set to \(0.8\), which is more strict than the common single threshold of \(0.5\) in COCO. In this work, the results are evaluated both with \(T=0.8\) and \(T=0.5\), which are denoted with \(AP^{\,FP}_{80}\) and \(AP^{\,FP}_{50}\), respectively. In the Data Anonymization Challenge, there is a mechanism of \texttt{Bad-Quality}: if an image is marked as \texttt{Bad-Quality}, its \(AP^{\,FP}_{80}\) is assigned with \(35\%\). We therefore record another two evaluation scores \(AP^{\,FP}_{80}*\,\) and \(AP^{\,FP}_{80}+\,\), where \(*\) denotes the \texttt{Bad-Quality} mechanism is used when calculating \(AP^{\,FP}_{80}\) and \(+\) denotes the images marked as \texttt{Bad-Quality} are excluded when calculating \(AP^{\,FP}_{80}\). 

\begin{figure}[h]
\begin{minipage}[b]{0.5\textwidth}
\begin{subfigure}[b]{\linewidth}
\centering
\includegraphics[width=0.7\linewidth]{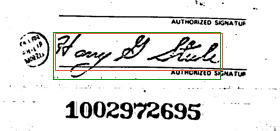}
\captionsetup{font=scriptsize}
\caption{\(GIoU=76.4;\; AP^{\,FP}_{80}=0;\; AP^{\,FP}_{50}=100 \)}
\end{subfigure}

\vspace*{2mm} 
\begin{subfigure}[b]{\linewidth}
\centering
\includegraphics[width=0.6\linewidth]{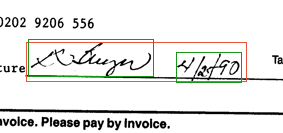}
\captionsetup{font=scriptsize}
\caption{\( GIoU=70.7;\; AP^{\,FP}_{80}=0;\; AP^{\,FP}_{50}=0\)}
\end{subfigure}
\end{minipage}
\begin{subfigure}[b]{0.45\textwidth}
\centering
\includegraphics[width=0.7\linewidth]{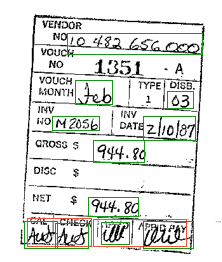}
\captionsetup{font=scriptsize}
\caption{\( GIoU=23.1;\; AP^{\,FP}_{80}=0;\; AP^{\,FP}_{50}=0\)}
\end{subfigure}

\vspace*{2mm} 
\caption{Comparison of different evaluation scores (in \%). \textit{Green}: ground-truth box; \textit{Red}: predicted box. See Section \ref{sec:eval} for the definitions of \( GIoU, AP^{\,FP}_{80}, AP^{\,FP}_{50}\). (a) shows the IoU threshold of 80\% might be too strict; (b) shows the \(AP\) family might not effectively indicate the decent quality of the prediction if the ground-truth contains multiple adjacent boxes and a larger one is detected, while \(GIoU\) could; (c) shows an unacceptable prediction where most of sensitive data are exposed, in which case the evaluation score of \(AP\) family (with a high threshold) is desired.}
\label{fig:eval}
\end{figure}

In addition, the overall evaluation score is calculated by averaging all image level scores, which applies to \(GIoU\) and \(AP^{\,FP}\). 
These 3 evaluation scores are illustrated with visual results of ground-truth and predicted bounding boxes in Fig.~\ref{fig:eval}. It is shown that there is no single evaluation score considered as \textit{silver bullet}, especially if the ground-truth annotations are not perfect. In this dataset, \textit{imperfect} is referred to as the fact that the neighboring handwritten regions are sometimes annotated as a single large box, sometimes as multiple separate small boxes. With this in mind, the results are evaluated with the following five scores: \(AP^{\,FP}_{80}\), \(AP^{\,FP}_{80}*\,\), \(AP^{\,FP}_{80}+\,\), \(AP^{\,FP}_{50}\) and \(GIoU\). Note that the first two evaluation scores (\(AP^{\,FP}_{80}\) and \(AP^{\,FP}_{80}*\,\)) are eligible in the SAP's Data Anonymization Challenge\footnotemark[1].
\clearpage

\subsection{Postprocessing}
The deep learning classifier outputs a confidence score for each corresponding class. This confidence score can be thresholded as a hyperparameter to control the false positive rate in the postprocessing step. It has been observed during the experiments that the best results (\textit{w.r.t.} \(AP^{\,FP}_{80}\)) can be achieved, if this confidence is thresholded as 0.8, which is thus chosen as default.

In addition, to reduce the overlapped bounding boxes, some postprocessing steps are applied in the end, which follows the simple criterion: one large box is preferable to multiple small ones. First, all the predicted bounding boxes
from each image are sorted in ascending order by their areas. Second, starting from the smallest one, each box is checked if the intersection area over the smaller box area is greater than a threshold (chosen as 0.9). If it is the case, the smaller box is omitted. This postprocessing is applied by default and we observed a minimal improvement, namely around 0.3\% in terms of \(AP^{\,FP}_{80}\).

\subsection{Results}\label{sec:res}
\begin{figure}[h]
\vspace{-0.3cm}
\centering
\begin{minipage}[t]{0.45\textwidth}
    \centering
    \includegraphics[width=4.6cm]{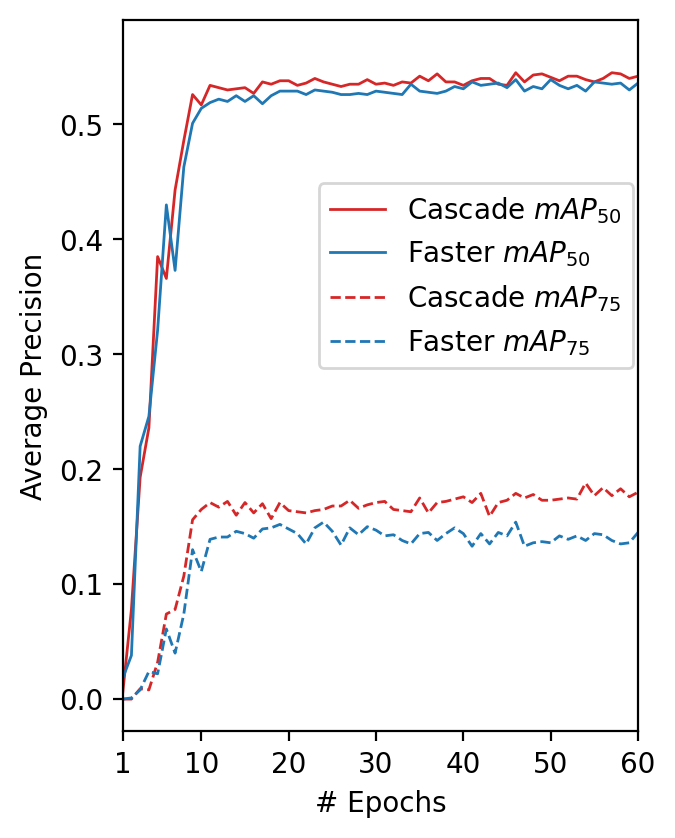}
    \captionof{figure}{\(mAP_{50}\) and \(mAP_{75}\) (\texttt{val}) of Cascade and Faster R-CNN. The cascade version surpasses Faster R-CNN by a larger margin under the more strict criterion. }
    \label{fig:margin}
\end{minipage}
\hspace{0.5cm}
\begin{minipage}[t]{0.45\textwidth}
    \centering
    \includegraphics[width=4.6cm]{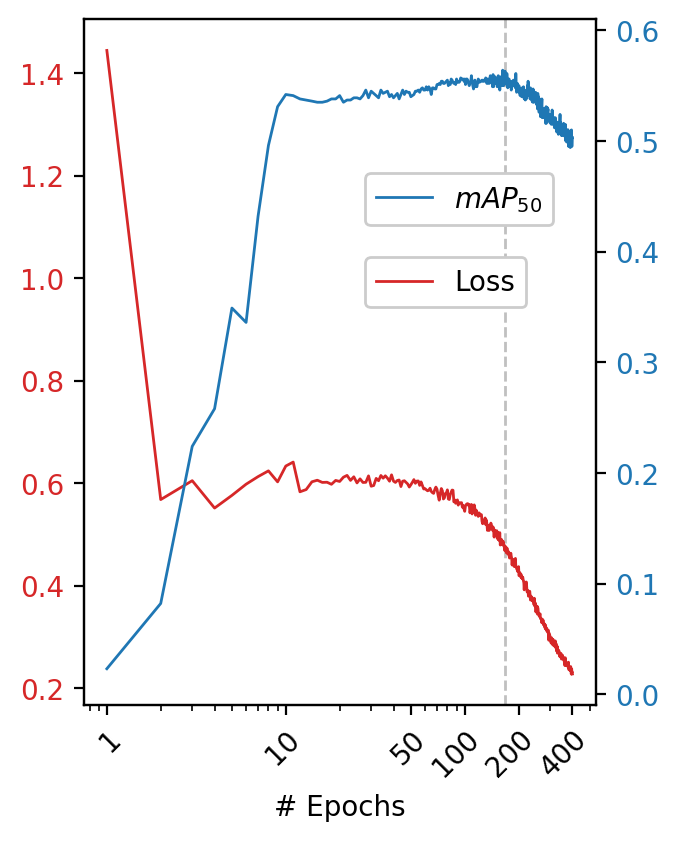}
    \captionof{figure}{Overall loss (bottom; left axis) and \(mAP_{50}\) (top; right axis) during the training. The dashed vertical line indicates the boundary of overfitting after around 170 epochs.}
    \label{fig:epoch}
\end{minipage}
\vspace{-0.3cm}
\end{figure}

In this section, the experimental results are presented regarding the different preprocessing steps (Table~\ref{tab:prep}), the influences of the mechanism of \texttt{Bad-Quality} (Table~\ref{tab:prep},~\ref{tab:lb}), the comparison of various deep learning networks (Table~\ref{tab:network}) and the results released on the leaderboard of Data Anonymization Challenge\footnotemark[1] (Table~\ref{tab:lb}). The improved performance from Fast R-CNN and Cascade R-CNN is specifically demonstrated (Fig.~\ref{fig:margin},~\ref{fig:epoch}). 
Furthermore, the examples of predicted handwritten regions
from the \texttt{val} set are visualized in Fig.~\ref{fig:vis}, together with the ground-truth bounding boxes.

\subsubsection{Preprocessing.} 
\begin{figure}[t]
\begin{minipage}[b]{\textwidth}
\centering
\captionof{table}{Handwriting detection results (in \%) using different deep learning networks. Inference speed is tested on a single RTX 2080 Ti GPU. Images which contain more than 3 detected boxes are marked as \texttt{Bad-Quality}. Dataset: \texttt{800train+198val+400test}. Column-wise best results are made bold.}
\label{tab:network}
  \begin{tabular}{lllllll}
    \toprule
    Network & Inference \;& \(AP^{\,FP}_{80}\) \;& \(AP^{\,FP}_{80}*\)\;& \(AP^{\,FP}_{80}+\) \;&\(AP^{\,FP}_{50}\) \;& \(GIoU\)\\
    \midrule
    YOLOv3 & \textbf{42} fps & 36.6 & 43.2 & 47.4 & 61.0 & 62.2\\
    RetinaNet & 11 fps & 27.9 & 40.8 &  44.4 & 51.7 & 54.9 \\
    Faster R-CNN & 11 fps & 37.1 &  45.3 &  57.2 & 62.1 & 66.6 \\
    Cascade R-CNN \;\;\;& 10 fps & \textbf{41.8} & \textbf{47.5} & \textbf{57.5} & \textbf{66.9} & \textbf{68.2} \\
  \bottomrule
\end{tabular}
\end{minipage}

\begin{minipage}[b]{1\linewidth}
  \begin{minipage}[b]{0.3\textwidth}
    \vspace{0.8cm}
    \centering
    \captionof{table}{Comparison with the leaderboard. \textit{OCR}: Tesseract with manual engineering. \textit{Service}: Google's Cloud API. \(\,^\star\) and \(\,^\dagger\) denote YOLOv3 results from the leaderboard and ours, respectively. \texttt{BQ}: if \texttt{Bad-Quality} is used.}
    \label{tab:lb}
  \begin{tabular}{lcc}
    \toprule
    Method & \(AP^{\,FP}_{80}\) & \texttt{BQ}\\
    \midrule
    YOLOv3\(\,^\star\) & 26.3 & \xmark\\
    OCR & 37.5 & \xmark \\
    Service & 42.5 & \xmark\\
    \midrule
    YOLOv3\(\,^\dagger\) & 36.6 & \xmark\\
    YOLOv3\(\,^\dagger\) & 43.2 & \cmark\\
    Cascade & 41.8 & \xmark\\
    Cascade & 47.5 & \cmark\\
  \bottomrule
\end{tabular}
\end{minipage}
\hfill
\begin{minipage}[t]{0.67\textwidth}
\centering
    \includegraphics[width=\textwidth]{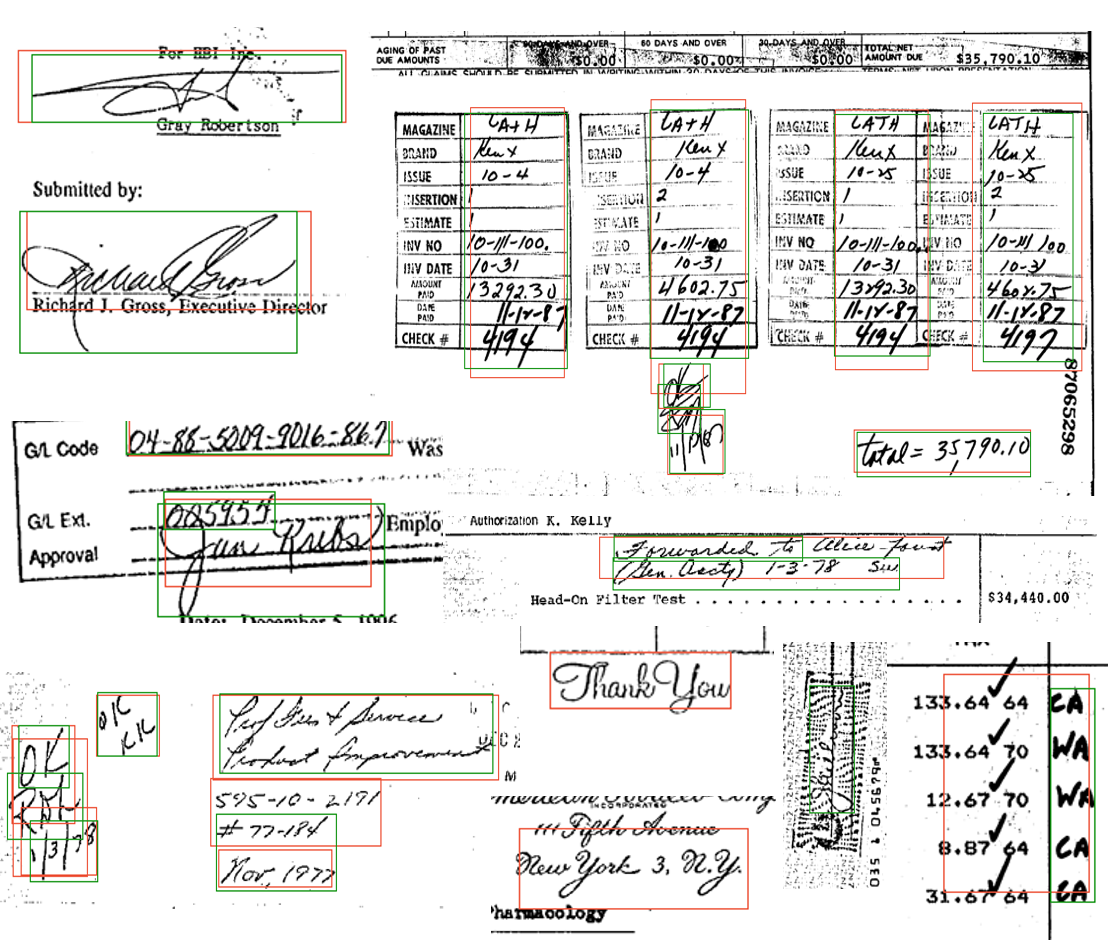}
    \captionof{figure}{Visual results of cropped handwritten regions from \texttt{val} set (Cascade R-CNN with \texttt{"o/o-/pre-"}). \textit{Green}: ground-truth box; \textit{Red}: predicted box.}
    \label{fig:vis}
\end{minipage}
\end{minipage}
\end{figure}

To begin with, different types of preprocessed images as the inputs of the deep learning networks are investigated based on Faster R-CNN and Cascade R-CNN. As shown in Table~\ref{tab:prep}, the preprocessed images alone can bring in worse results, compared to the original ones, while the concatenated preprocessed and original images as inputs (\texttt{"o/o-/pre-"}) have achieved the best result \textit{w.r.t. \(AP^{\,FP}_{80}\)}, which are therefore used as default in the following experiments (Table~\ref{tab:network},~\ref{tab:lb}). It is believed that the fused preprocessed images could be served as an \textit{attention} mechanism, which highlights the regions more likely containing the desired objects. Moreover, it is empirically shown in Table~\ref{tab:prep} that sparse inputs have not always yielded better results, if compared the negated inputs with the original ones.

\begin{figure}[h]
\begin{minipage}[b]{\textwidth}
\centering
  \captionof{table}{Handwriting detection results (in \%) with different preprocessing steps as inputs. Images which contain more than 3 detected boxes are marked as \texttt{Bad-Quality}. Dataset: \texttt{600train+198val+200test}. Input abbreviations see Section~\ref{sec:prep}. Column-wise best results are made bold.}
  \label{tab:prep}
\begin{tabular}{l l lllll}
    \toprule
    Network & Input & \(AP^{\,FP}_{80}\)  \;& \(AP^{\,FP}_{80}*\) \;& \(AP^{\,FP}_{80}+\)  \;&\(AP^{\,FP}_{50}\)  \;& \(GIoU\)\\
    \midrule
    Faster R-CNN & \texttt{o}  & 34.2 & 45.4 & 59.2 & 65.5 & 64.6 \\
    Faster R-CNN   & \texttt{o-} & 35.1 & 45.7 & 58.2 & 64.9 & 65.1 \\
    Faster R-CNN & \texttt{pre}  & 31.3 & 43.1 & 55.9 & 54.0 & 56.3 \\
    Faster R-CNN  &  \texttt{pre-} & 29.6 & 42.4 & 54.0 & 54.6 & 55.3 \\
    Faster R-CNN   & \texttt{o/pre} & 34.6 & 43.9 & 56.3 & 64.1 & 64.4 \\
    Faster R-CNN & \texttt{o-/pre-}  & 35.1 & 44.5 & 53.3 & \textbf{66.7} & 65.4 \\
    Faster R-CNN & \texttt{o/o-/pre}  & 37.2 & 45.6 & 59.6 & 63.3 & 64.9 \\
    Faster R-CNN  & \texttt{o/o-/pre-}\;\;\;\;   & 35.1 & 45.0 & 57.4 & 62.4 & 64.3 \\
    \midrule
    Cascade R-CNN   \;\;\; & \texttt{o}  & 37.7 & 45.7 & 56.6 & 65.6 & 66.4 \\
    Cascade R-CNN   & \texttt{o-} & 34.7 & 42.9 & 49.2 & 65.1 & 66.5 \\
    Cascade R-CNN  &  \texttt{pre} & 31.9 & 41.6 & 47.5 & 55.4 & 56.7 \\
    Cascade R-CNN   & \texttt{pre-}  & 32.2 & 42.0 & 47.6 & 57.0 & 58.7 \\
    Cascade R-CNN   & \texttt{o/pre} & 36.3 & 44.3 & 56.0 & 64.4 & 66.4 \\
    Cascade R-CNN & \texttt{o-/pre-}  & 35.0 & 44.7 & 56.6 & 64.1  & 64.3 \\
    Cascade R-CNN & \texttt{o/o-/pre}  & 37.2 &\textbf{46.9} & \textbf{60.0} & 64.4 & 66.3 \\
    Cascade R-CNN  & \texttt{o/o-/pre-}  & \textbf{38.3} & 46.0 & 56.5 & 65.0 & \textbf{66.8} \\
  \bottomrule
\end{tabular}
\end{minipage}
    \vspace{-0.5cm}
\end{figure}

\subsubsection{Mechanism of \texttt{Bad-Quality}.} As introduced in Section~\ref{sec:eval}, the mechanism of \texttt{Bad-Quality} is provided in the challenge\footnotemark[1], which has been made full use of with the evaluation scores \(AP^{\,FP}_{80}*\) and \(AP^{\,FP}_{80}+\,\). It is observed based on the \texttt{val} set that the performance tends to be worse with the increasing number of the detected bounding boxes. With this in mind, all images with more than 3 detected bounding boxes are marked as \texttt{Bad-Quality}. The purpose of this mechanism is to raise the abnormal or complicated cases and turn to the manual process, which is practical in the industry. However, if the number of images marked as \texttt{Bad-Quality} is too large to maintain the productive advantage of machines, the evaluation scores \(AP^{\,FP}_{80}*\) and \(AP^{\,FP}_{80}+\) might bring in pseudo good results. Therefore, the number of images marked as \texttt{Bad-Quality} is loosely limited up to 50\% of all images. Conclusively, the mechanism of \texttt{Bad-Quality} is believed to be a flexible trick to deal with the hard cases.

\subsubsection{Faster and Cascade R-CNN.}
Not surprisingly, it is shown in Table~\ref{tab:prep} that Cascade R-CNN outperforms Faster R-CNN in general except for \(AP^{\,FP}_{50}\), as Cascade R-CNN focuses on the object detection of higher quality and might perform worse than Faster R-CNN in terms of \(AP^{\,FP}_{50}\). As shown in Fig.~\ref{fig:margin}, the cascade version surpassed Faster R-CNN by a larger margin under the more strict criterion, when compared \(mAP_{75}\) to \(mAP_{50}\) (using COCO's evaluation scores~\cite{lin2014coco} for brevity) in the \texttt{val} set. In addition, as depicted in Fig.~\ref{fig:epoch}, the training progress has been overfitted after around 170 epochs, from where the loss values and \(mAP_{50}\) start decreasing. Thus, the chosen 200 epochs during all the experiments are appropriate.

\subsubsection{Comparison with other approaches.}
 Some of the popular deep learning networks for object detection are compared in Table~\ref{tab:network}, including YOLOv3~\cite{redmon2018yolov3}, RetinaNet~\cite{lin2017focal}, Faster R-CNN~\cite{ren2016faster} and Cascade R-CNN~\cite{cai2019cascade}. They are implemented with~\cite{mmdetection}. Darknet-53~\cite{darknet13} is used as the backbone for YOLOv3 and ResNeXt-101~\cite{Xie2016resnext} for the other three. The dataset used in this experiment, \texttt{800train+198val+400test} (see Section~\ref{sec:dataset}), is identical with the one used in the leaderboard. The images are preprocessed to the form of \texttt{"o/o-/pre-"} as described in Section~\ref{sec:prep}. The results show that Cascade R-CNN outperforms other networks, with the trade-off regarding the inference speed on a single RTX 2080 Ti GPU due to the extra computational overhead though. As showcased in Table~\ref{tab:lb}, the results of our approaches are compared with those submitted to the leaderboard. The first three rows are the results from the leaderboard, and the following four rows are our approaches. Our best achieved result (with Cascade R-CNN and \texttt{BQ}) has surpasses previous submissions on the leaderboard. Without considering the mechanism of \texttt{BQ}, however, the paid Google AutoML Vision API is slightly more advantageous (by 0.7\%). Besides, it is noticeable that our YOLOv3 result (implemented by~\cite{mmdetection}) has outperformed the version submitted to the leaderboard (implemented by~\cite{ImageAI}) by more than 10\% in terms of \(AP^{\,FP}_{80}\). 
  
\subsection{Generalizability}
English is the vast majority of the languages used in the dataset. Other languages such as Dutch or German are also included. However, the deep learning network is not expected to recognize the discrepancy of different languages. It is natural to categorize the languages using Latin alphabets indiscriminately. In this section, it is tested if the trained model works on the redacted real-world images in \textit{foreign} languages.

\begin{figure}[h]
    \centering
    \includegraphics[width=\textwidth]{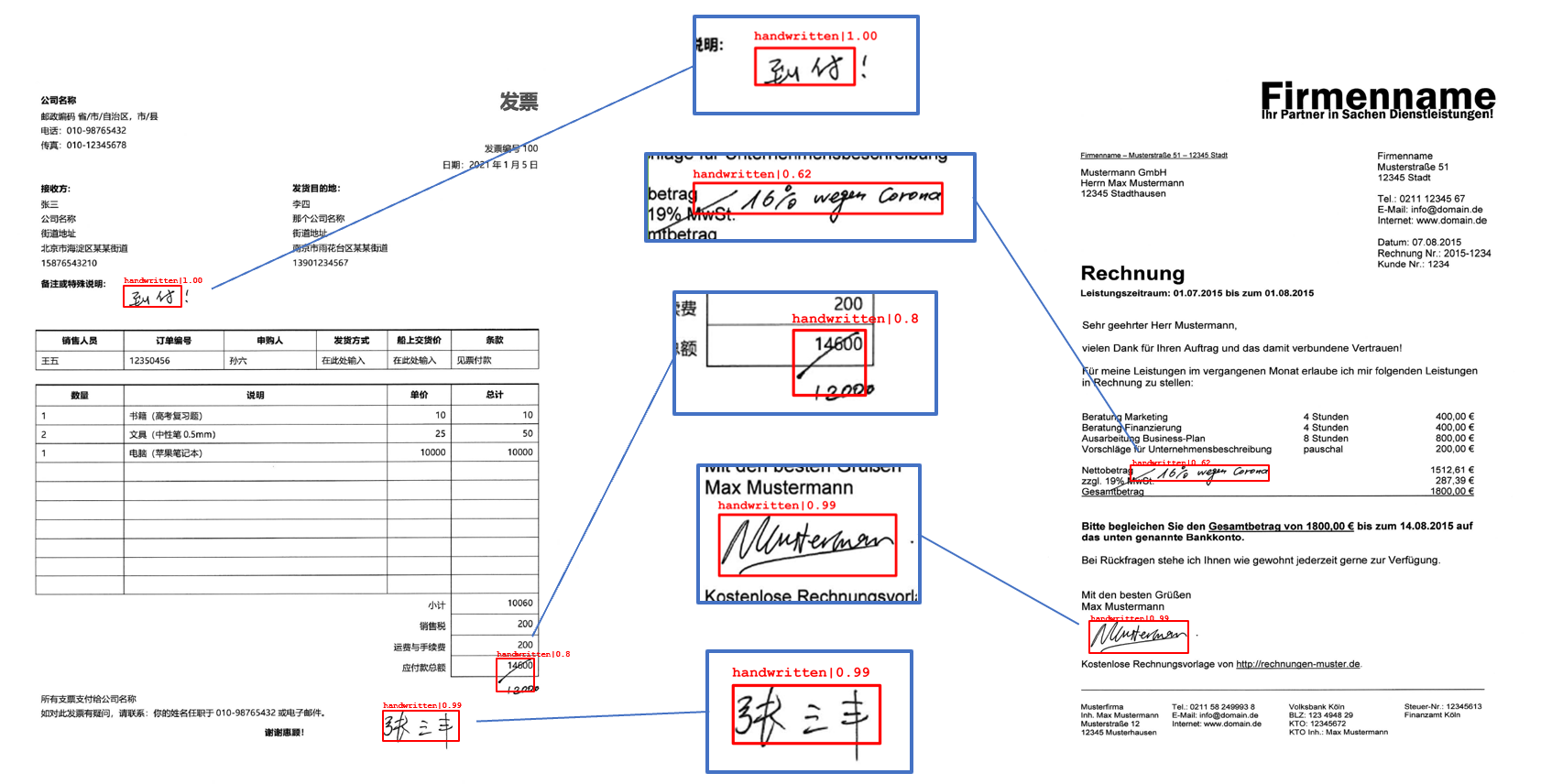}
    \caption{Test of generalizability on toy examples with fictitious and unseen invoices. The handwritten regions are accurately localized from the images in Chinese and German.}
    \label{fig:cnde}
    \vspace{-0.8cm}
\end{figure}

Fig.~\ref{fig:cnde} illustrates two toy examples of fictitious and unseen invoices to evaluate the generalizability of the trained model. The model used to localize the handwritten regions is Cascade R-CNN. It is noteworthy that the language in the left image in Fig.~\ref{fig:cnde} is Chinese, which can be considered as a foreign language in the dataset. Analogous to the German invoice demonstrated in the right one, the handwritten regions of both images are accurately detected as desired. The generalizability of the R-CNN family has also been observed by~\cite{zhong2017improved} during the text
detection in natural scene images.

It is believed to be beneficial in the industry, if the model can be trained once and applicable to various cases. Additionally, it has also raised the common question of what the deep learning network has learned. In this case, it is supposed that the \textit{irregularity} might be learned to discriminate the printed and handwritten texts.
\clearpage

\section{Discussion}
\subsection{Conclusion}
In this work, we present an object detection based approach to localize the handwritten regions, which is effective, fast and applicable to the unseen languages. First, the influences of the preprocessing steps are investigated. It has been empirically found that the fused concatenation of original and preprocessed images as the inputs can achieve the best performance. Second, different deep learning networks are compared. It is noticeable that the improvement from Faster R-CNN to Cascade R-CNN can be reproduced and the \textit{high quality} characteristic of the cascade version suits the problem of the handwriting localization well. The results of our approaches can be served as a baseline of deep learning approaches in the handwriting localization problem. At last, the generalizability of the deep learning approach is impressive. The learned model is capable to successfully detect the handwritten regions on the real-world unseen images, even for those in the unseen language of Chinese. We believe it is of great interest both for the future research and for the industrial applications.

\subsection{Outlook}
As showcased in Fig.~\ref{fig:vis}, some printed cursive texts are also detected as handwriting. It remains challenging to distinguish such nuanced discrepancies. Furthermore, apart from the object detection approaches, other proposals in the field of computer vision can also be adopted to differ the handwritten texts from the printed ones. One example is the anomaly detection, where the printed texts can be considered as the \textit{normal} instances, since they are more regularly shaped. Thanks to the algorithms like \textit{variational autoencoder} (VAE)~\cite{kingma2013auto}, it is also promising to accomplish such tasks in a semi-supervised or even unsupervised manner. The other benefit of using the algorithms like VAE is that the learned intermediate representations can also be exploited to synthesize the artificial signatures, further enhancing the anonymization without eliminating the existence of such entities. 

%
%
\bibliographystyle{splncs04}
\bibliography{bib}

\begin{thebibliography}{10}
\providecommand{\url}[1]{\texttt{#1}}
\providecommand{\urlprefix}{URL }
\providecommand{\doi}[1]{https://doi.org/#1}

\bibitem{cai2019cascade}
Cai, Z., Vasconcelos, N.: Cascade r-cnn: high quality object detection and
  instance segmentation. IEEE Transactions on Pattern Analysis and Machine
  Intelligence  (2019)

\bibitem{canny1986computational}
Canny, J.: A computational approach to edge detection. IEEE Transactions on
  pattern analysis and machine intelligence (6),  679--698 (1986)

\bibitem{mmdetection}
Chen, K., Wang, J., Pang, J., Cao, Y., Xiong, Y., Li, X., Sun, S., Feng, W.,
  Liu, Z., Xu, J., Zhang, Z., Cheng, D., Zhu, C., Cheng, T., Zhao, Q., Li, B.,
  Lu, X., Zhu, R., Wu, Y., Dai, J., Wang, J., Shi, J., Ouyang, W., Loy, C.C.,
  Lin, D.: {MMDetection}: Open mmlab detection toolbox and benchmark. arXiv
  preprint arXiv:1906.07155  (2019)

\bibitem{duda1972hough}
Duda, R.O., Hart, P.E.: Use of the hough transformation to detect lines and
  curves in pictures. Communications of the ACM  \textbf{15}(1),  11--15 (1972)

\bibitem{voc}
Everingham, M., Eslami, S.M.A., Van~Gool, L., Williams, C.K.I., Winn, J.,
  Zisserman, A.: The pascal visual object classes challenge: A retrospective.
  International Journal of Computer Vision  \textbf{111}(1),  98--136 (Jan
  2015)

\bibitem{gidaris2016locnet}
Gidaris, S., Komodakis, N.: Locnet: Improving localization accuracy for object
  detection. In: Proceedings of the IEEE conference on computer vision and
  pattern recognition. pp. 789--798 (2016)

\bibitem{girshick2015fast}
Girshick, R.: Fast r-cnn. In: Proceedings of the IEEE international conference
  on computer vision. pp. 1440--1448 (2015)

\bibitem{harley2015evaluation}
Harley, A.W., Ufkes, A., Derpanis, K.G.: Evaluation of deep convolutional nets
  for document image classification and retrieval. In: 2015 13th International
  Conference on Document Analysis and Recognition (ICDAR). pp. 991--995. IEEE
  (2015)

\bibitem{he2017mask}
He, K., Gkioxari, G., Doll{\'a}r, P., Girshick, R.: Mask r-cnn. In: Proceedings
  of the IEEE international conference on computer vision. pp. 2961--2969
  (2017)

\bibitem{kingma2013auto}
Kingma, D.P., Welling, M.: Auto-encoding variational bayes. arXiv preprint
  arXiv:1312.6114  (2013)

\bibitem{lecun1989backpropagation}
LeCun, Y., Boser, B., Denker, J.S., Henderson, D., Howard, R.E., Hubbard, W.,
  Jackel, L.D.: Backpropagation applied to handwritten zip code recognition.
  Neural computation  \textbf{1}(4),  541--551 (1989)

\bibitem{lin2017focal}
Lin, T.Y., Goyal, P., Girshick, R., He, K., Doll{\'a}r, P.: Focal loss for
  dense object detection. In: Proceedings of the IEEE international conference
  on computer vision. pp. 2980--2988 (2017)

\bibitem{lin2014coco}
Lin, T.Y., Maire, M., Belongie, S., Hays, J., Perona, P., Ramanan, D.,
  Doll{\'a}r, P., Zitnick, C.L.: Microsoft coco: Common objects in context. In:
  European conference on computer vision. pp. 740--755. Springer (2014)

\bibitem{liu2016ssd}
Liu, W., Anguelov, D., Erhan, D., Szegedy, C., Reed, S., Fu, C.Y., Berg, A.C.:
  Ssd: Single shot multibox detector. In: European conference on computer
  vision. pp. 21--37. Springer (2016)

\bibitem{malik2013icdar}
Malik, M.I., Liwicki, M., Alewijnse, L., Ohyama, W., Blumenstein, M., Found,
  B.: Icdar 2013 competitions on signature verification and writer
  identification for on-and offline skilled forgeries (sigwicomp 2013). In:
  2013 12th International Conference on Document Analysis and Recognition. pp.
  1477--1483. IEEE (2013)

\bibitem{marti2002iam}
Marti, U.V., Bunke, H.: The iam-database: an english sentence database for
  offline handwriting recognition. International Journal on Document Analysis
  and Recognition  \textbf{5}(1),  39--46 (2002)

\bibitem{ImageAI}
Moses, Olafenwa, J.: Imageai, an open source python library built to empower
  developers to build applications and systems with self-contained computer
  vision capabilities (mar 2018--),
  \url{https://github.com/OlafenwaMoses/ImageAI}

\bibitem{nayef2019icdar2019}
Nayef, N., Patel, Y., Busta, M., Chowdhury, P.N., Karatzas, D., Khlif, W.,
  Matas, J., Pal, U., Burie, J.C., Liu, C.l., et~al.: Icdar2019 robust reading
  challenge on multi-lingual scene text detection and
  recognition—rrc-mlt-2019. In: 2019 International Conference on Document
  Analysis and Recognition (ICDAR). pp. 1582--1587. IEEE (2019)

\bibitem{ortega2003mcyt}
Ortega-Garcia, J., Fierrez-Aguilar, J., Simon, D., Gonzalez, J., Faundez-Zanuy,
  M., Espinosa, V., Satue, A., Hernaez, I., Igarza, J.J., Vivaracho, C.,
  et~al.: Mcyt baseline corpus: a bimodal biometric database. IEE
  Proceedings-Vision, Image and Signal Processing  \textbf{150}(6),  395--401
  (2003)

\bibitem{darknet13}
Redmon, J.: Darknet: Open source neural networks in c.
  \url{http://pjreddie.com/darknet/} (2013--2016)

\bibitem{redmon2016you}
Redmon, J., Divvala, S., Girshick, R., Farhadi, A.: You only look once:
  Unified, real-time object detection. In: Proceedings of the IEEE conference
  on computer vision and pattern recognition. pp. 779--788 (2016)

\bibitem{redmon2018yolov3}
Redmon, J., Farhadi, A.: Yolov3: An incremental improvement. arXiv preprint
  arXiv:1804.02767  (2018)

\bibitem{ren2016faster}
Ren, S., He, K., Girshick, R., Sun, J.: Faster r-cnn: Towards real-time object
  detection with region proposal networks. IEEE transactions on pattern
  analysis and machine intelligence  \textbf{39}(6),  1137--1149 (2016)

\bibitem{smith2007overview}
Smith, R.: An overview of the tesseract ocr engine. In: Ninth international
  conference on document analysis and recognition (ICDAR 2007). vol.~2, pp.
  629--633. IEEE (2007)

\bibitem{Xie2016resnext}
Xie, S., Girshick, R., Dollár, P., Tu, Z., He, K.: Aggregated residual
  transformations for deep neural networks. arXiv preprint arXiv:1611.05431
  (2016)

\bibitem{zhong2017improved}
Zhong, Z., Sun, L., Huo, Q.: Improved localization accuracy by locnet for
  faster r-cnn based text detection. In: 2017 14th IAPR International
  Conference on Document Analysis and Recognition (ICDAR). vol.~1, pp.
  923--928. IEEE (2017)

\end{thebibliography}
\end{document}